\documentclass[sn-mathphys-num]{sn-jnl}

\usepackage{graphicx}%
\usepackage{multirow}%
\usepackage{amsmath,amssymb,amsfonts}%
\usepackage{amsthm}%
\usepackage{mathrsfs}%
\usepackage[title]{appendix}%
\usepackage{xcolor}%
\usepackage{textcomp}%
\usepackage{manyfoot}%
\usepackage{booktabs}%
\usepackage{algorithm}%
\usepackage{algorithmicx}%
\usepackage{algpseudocode}%
\usepackage{listings}%
\usepackage{subcaption}

\theoremstyle{thmstyleone}%
\theoremstyle{thmstyletwo}%

\theoremstyle{thmstylethree}%

\begin{document}

\title[]{Segmentation, Classification and Interpretation of Breast Cancer Medical Images using Human-in-the-Loop Machine Learning}


\author[1]{\fnm{David} \sur{V\'azquez-Lema}}\email{david.vazquez7@udc.es}

\author*[1]{\fnm{Eduardo} \sur{Mosqueira-Rey}}\email{eduardo@udc.es}

\author[1]{\fnm{Elena} \sur{Hern\'andez-Pereira}}\email{elena.hernandez@udc.es}

\author[1]{\fnm{Carlos} \sur{Fern\'andez-Lozano}}\email{carlos.fernandez@udc.es}

\author[1]{\fnm{Fernando} \sur{Seara-Romera}}\email{f.searar@udc.es}

\author[2]{\fnm{Jorge} \sur{Pombo-Otero}}\email{jorge.pombo.otero@sergas.es}

\affil*[1]{\orgdiv{Department of Computer Science and Information Technologies}, \orgname{University of Coruña (CITIC)}, \orgaddress{\street{Rúa Maestranza, 9}, \city{La Coruña}, \postcode{15001}, \state{Galicia}, \country{Spain}}}

\affil*[2]{\orgdiv{Servicio de Anatomía Patológica}, \orgname{Complejo Hospitalario Universitario de A Coruña (CHUAC)}, \orgaddress{\street{As Xubias, 84}, \city{A Coru\~na}, \postcode{15006}, \country{Spain}}}


\abstract{
This paper explores the application of Human-in-the-Loop (HITL) strategies in training machine learning models in the medical domain. In this case a ‘doctor-in-the-loop’ approach is proposed to leverage human expertise in dealing with large and complex data. Specifically, the paper deals with the integration of genomic data and Whole Slide Imaging (WSI) analysis of breast cancer. Three different tasks were developed: segmentation of histopathological images, classification of this images regarding the genomic subtype of the cancer and, finally, interpretation of the machine learning results. The involvement of a pathologist helped us to develop a better segmentation model and to enhance the explainatory capabilities of the models, but the classification results were suboptimal, highlighting the limitations of this approach: despite involving human experts, complex domains can still pose challenges, and a HITL approach may not always be effective.
}




\maketitle

\section{Introduction}\label{sec:introduction}

Cancer is a highly heterogeneous disease and a major contributor to global mortality, responsible for about 1 in every 6 deaths \cite{Siegel2024}. Cancer cells are often undifferentiated, complicating their characterization, and exhibit a high capacity for proliferation/migration within the tumor microenvironment. Different subtypes are often identified for each specific type of cancer, with varying prognoses, survival times, and responses to treatments.

For women, breast cancer (BC) has surpassed lung cancer as the most diagnosed cancer worldwide accounting for 32\% of cases \cite{Siegel2024,Dizon2024}, and 23\% of mortality. Overall, breast cancer, lung cancer, and colorectal cancer account for 51\% of all new diagnoses annually for women. The incidence rates of female breast cancer have experienced a gradual rise of approximately 0.6\% annually since the mid-2000s, primarily attributed to the increased detection of localized-stage and hormone receptor-positive cases \cite{Giaquinto2022}.

Currently, the diagnosis of BC in daily clinical practice is made by immunohistochemical analysis of cancerous breast tissue removed during surgery, studying the presence or absence of estrogen receptor (ER), progesterone receptor (PR), HER2 membrane protein, and the Ki-67 proliferation index. In those healthcare centers with the capacity for massive sequencing, the most common approach is the study of the tumor cell transcriptome based on the expression of 50-gene signature \cite{Parker2008} in paraffin-embedded tumor tissue, also called intrinsic subtypes (IS). There is no perfect correlation between immunohistochemical and genomic classification \cite{Pascual2019}, with the latter being more accurate but much more expensive and less common in daily clinical practice. 

As mentioned earlier, in daily clinical practice in hospitals, and despite the rapid advances in genomic medicine, pathologists still carry out patient stratification through manual review of histopathology slides with a light microscope. Images are the product of molecular mechanisms occurring in tissues at the molecular level. The shape and distribution of the different visible elements in pathological images are used to determine the type of cell, tissue, and cancer, as well as its immunohistochemical response, and are crucial for the identification, classification, and prognosis of cancer \cite{Gurcan2009}. However, this review is often subjective, time-consuming, and there are significant intra- and inter-operator variations, which often lead to discrepancies in diagnostic and subtyping results and sometimes even misdiagnosis. However, histopathology images remain to this day the fundamental method for the clinical diagnosis of tumor progression.

Both the tumor genome analysis approach and the histopathological image analysis approach are complementary and provide i) information about the genetic origin/component of the disease, ii) spatial information (shape, distribution, and presence of different cell types) derived from the previous process, and iii) interaction within its tumor microenvironment.

In this paper we present a comprehensive approach to histological image analysis of breast cancer, focusing on segmentation and classification tasks related to cancer types according to their genetic information. In addition, we introduce interpretability measures to shed light on model performance during classification.

Throughout the different phases of our research, we actively involved a pathologist in a human-in-the-loop process \cite{mosqueira2023human}. This collaborative effort was intended to improve the quality of the results by incorporating expert knowledge. Let us delve into the details of our collaboration.

Our first step involved automated segmentation of histological slides prepared from formalin-fixed tissue in order to simplify the information contained in those images and to make them easier to analyze. This segmentation will try to annotate the different tissue components using different colors to be used in further steps. The pathologist’s role was crucial here as he meticulously reviewed the segmentation results and provided corrections where necessary. His expertise ensured accurate delineation of relevant structures within the images, usually trying to generalize areas into more uniform groups.

The second step was to proceed with the classification of the segmentation result for the different images, based on cancer types. Unfortunately, the problem that we were facing here were rather complex and, with the available data the results of the different classification techniques applied were not optimal. The pathologist guided us by identifying critical areas that the model should focus on during classification but this information was not helpful to improve the classification results. 

Rather than discarding the pathologist’s input, we decided to reformulate it for interpretability and also as a debugging process. The pathologist evaluated the interpretability results of the system and we used this information to see why the models failed and as feedback for our hyperparameter optimization process. Our goal was not only select the best-performing model but also the model that offers the best explainability, something that is very relevant in medical domains.

Thus, our final contribution has been to demonstrate how a human-in-the-loop process allowed us to improve certain aspects of machine learning model building and to remark some of the limitations of this type of process: aside from the complexities of involving human experts we have seen that, in complex domains, even human experts find it difficult to solve certain problems, and their collaboration may not always be helpful.

The paper is structured as follows: In section \ref{sec:state_art} we review the current state of the art on the subject of breast cancer, human-in-the-loop ML techniques and ML model explainability algorithms. Section \ref{sec:segmentation} describes the segmentation algorithm an the corrections made by the pathologist. Section \ref{sec:classification} describes the classification models developed and their problems. Section \ref{sec:interpretation} describes the methods used to obtain information on the internal functioning of the model, trying to explain the decisions taken and how the pathologist's feedback on these data help us to develop more interpretable models. We ultimately concluded with our discussion and conclusions about the matter in section \ref{sec:discussion}.

\section{State of the Art}\label{sec:state_art}

\subsection{Context of the problem}\label{sec:contet}

The diagnosis of complex diseases, such as cancer, has long relied on Whole-Slide Histopathological Images (WSIs) as a clinical gold standard tool \cite{Kosaraju2022,Bera2019}. This is primarily due to the fact that diagnosis is predominantly based on morphological patterns observed in WSIs. Such images arise from the need to analyze histopathological sections of biopsies in an automated manner, aiming to reduce the highly costly manual process and address the lack of quantitative precision in cellular characteristics. Before WSI automated analysis, a prevalent method for diagnosing cancer involves examining tumor tissue sections under a microscope, which are typically stained with hematoxylin and eosin (H\&E) dyes \cite{Su2022-pd,Rosai2007-ru}. This staining process enables pathologists to qualitatively evaluate cancer types, stages, and approximate tumor purity based on the H\&E-stained image of a biopsy section. Moreover, histopathologic examination often reveals various cell types, organic conditions, and/or cellular locations within intricate tissues. However, there is limited agreement among pathologists regarding diagnoses \cite{Van_der_Laak2021-fa}.

Deep learning is the most popular method for analyzing WSI, particularly for tumor classification \cite{Bera2019-uz}. However, a major challenge for deep learning is the need for a large set of accurately labeled data. Many techniques require WSI that have been annotated by pathologists. This process of creating a training dataset can be labor-intensive, especially due to the differences in interpretation among pathologists. The novelty of the proposed work lies in integrating human involvement into the training process, thereby adopting a human-in-the-loop (HITL) approach. To the best of our knowledge, our study is the first to apply HITL deep-learning method for integration of genomic data and WSI-based analysis. According to a recent systematic review \cite{SCHNEIDER202280}, the utilization of deep learning techniques for WSI analysis in combination with omics data remains limited, with a focus primarily on survival analysis. This indicates an open avenue for further research in the field.

This study will focus on a specific type of cancer, namely, breast cancer. Therefore, it is necessary to understand the known genomic differences that characterize this pathology. Breast cancer intrinsic molecular subtypes represent unique biological entities, each characterized by specific gene expression profiles. These widely accepted subtypes, including Luminal A, Luminal B, HER2-Enriched, Normal-like, and Basal-like, hold prognostic value and offer potential therapeutic insights \cite{Schettini2022}. Luminal A cancers are characterized by low proliferation markers and the presence of hormone receptor-positive genes. They grow slowly and have a good prognosis with hormone therapy. In contrast, Luminal B tumors have higher hormone receptor expression and grow faster with a worse prognosis requiring more intensive treatment. HER2-enhanced tumors overexpress the HER2 gene and are more aggressive with rapid growth compared to luminal tumors. Basal-like tumors are characterized by the absence of ER/PR hormone receptors and HER2 amplification. They are known as ``triple-negative'' breast cancers, with aggressive growth and chemotherapy as the main treatment. Normal-like cancers share gene expression patterns with normal breast tissue and are often excluded from analyses due to their similarity to non-cancerous tissue.

\subsection{Dataset}\label{sec:dataset}

The Cancer Genome Atlas (TCGA) \cite{CancerGenomeAtlasResearchNetwork:2013} is a public and collaborative database focused in cancer research that aimed to study 33 different types of cancer, such as breast, lung or blood, among others. The TCGA project started in 2006 with the joint effort between the National Cancer Institute(NCI) and the National Human Genome Research. Since then, information about over 11.000 patients is publicly available, containing genomic, epigenomic, transcriptomic and proteomic data. 
The Cancer Imaging Archive (TCIA) \cite{TCIA:2013} is a open access repository that hosts a large archive of medical images of various types of cancer. All the images provided can be downloaded, facilitating the research of the various cancers. 
TCGA and TCIA projects are focused on different aspects of the cancer research, but both are related. In this project, this relationship will allow to download the images belonging to the patients included in TCGA, allowing the creation of a custom dataset. 

Our initial dataset was composed by a custom set of patients downloaded from TCGA, containing personal, clinical and genomic information. The most important features for our project are the unique patient identifier, that identify all the tissue images related with them, and the cancer type. 

This project is centered solely in breast cancer. The dataset was custom-designed, composed by a total of 1230 patients obtained from the TCGA database. Over the 10\% of the dataset did not contain cancer type information, a significant amount of missing values that needed to be clear. This patients are not useful for the project and were immediately discarded. 

\begin{figure}[htbp!]
    \centering
    \begin{subfigure}{0.45\textwidth}
        \includegraphics[width=\textwidth]{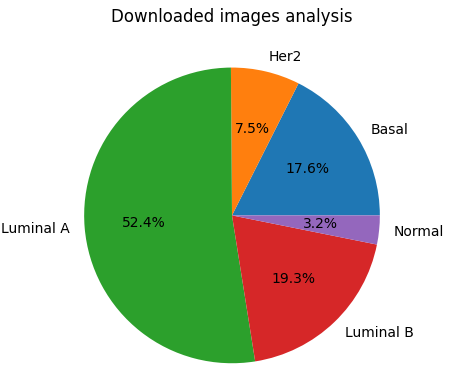}
        \caption{}
        \label{fig:downloaded_images_analysis}
    \end{subfigure}
    \hfill
    \begin{subfigure}{0.45\textwidth}
        \includegraphics[width=\textwidth]{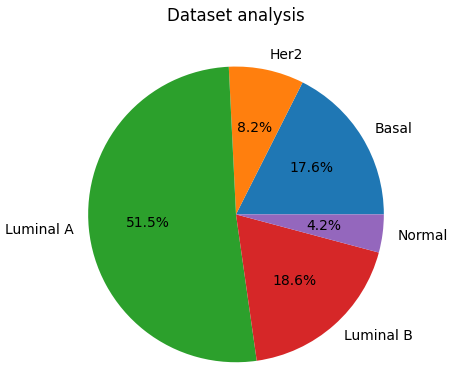}
        \caption{}
        \label{fig:dataset_analysis}
    \end{subfigure}
    \caption{(a) Percentage of downloaded images belonging to each cancer type (b)  Percentage of the final dataset images belonging to each cancer type.}
    \label{fig:data_analysis}
\end{figure}

After cleaning the missing values, a total of 2783 images related to the remaining patient identifiers were downloaded from the TCIA repository. The images downloaded use a digital and high resolution representation of the microscope slides. 
The percentage of images related with each cancer type is shown in Figure \ref{fig:downloaded_images_analysis}. 

The remaining images presented at least two tissues in each image or color pen marks that needed to be removed, so some preprocessing task was needed. Also, the images need to be converted to a more manageable format to apply all the previously mentioned preprocessing. 

Finally, the images that didn't contain enough tissue after the preprocessing were discarded. The final built dataset is composed by over 1000 tissue images and its distribution among cancer types is shown in Figure \ref{fig:dataset_analysis}. The distribution shows that the dataset is imbalanced. Almost half of the final cases belong to the Luminal A class, followed by the Basal and Luminal B classes with over the 15\% of the data each. Her2 and Normal classes are under-represented and will be discarded in some of the experiments.

To increase the amount of data available, some data augmentation operations have been applied: horizontal flipping, vertical flipping and zooming.

\subsection{Human-in-the-loop Machine Learning}\label{sec:hitl-ml}

Human-in-the-Loop Machine Learning (HITL-ML) is the practice of uniting human and machine intelligence to create effective machine learning algorithms \cite{mosqueira2023human}. In this approach, humans can play several roles before, during and after the learning process with the goal of enhancing the capabilities of ML models by incorporating human intelligence, judgment, and feedback.

\begin{figure*}[htbp!]
    \centerline{\includegraphics[width=\textwidth]{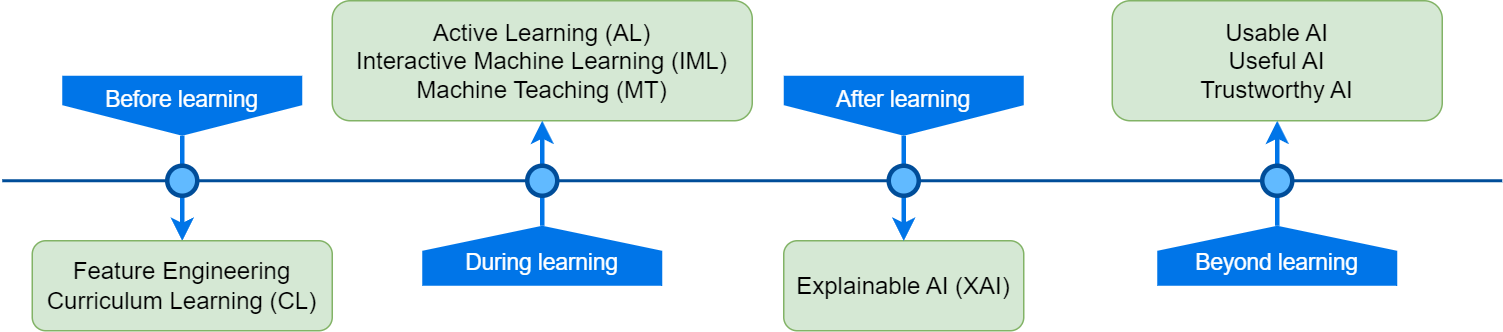}}
    \caption{Classification of HITL methods.}
    \label{fig:hitl_classification}
\end{figure*}

The main HITL methods revolve around the learning process, but have also been proven useful for tasks performed just before or after the learning algorithm. Taking this into account we can classify HITL methods according to their relationship to the learning process \cite{mosqueira2023addressing}, as depicted in Fig. \ref{fig:hitl_classification}.

\textbf{Before learning}, humans play a critical role in tasks such as \textit{Feature Engineering}, even in the era of Deep Learning. This is because humans possess domain expertise and intuition about the data they work with and they can uncover hidden patterns and create meaningful features and decide how to transform raw data into numeric representations that can feed the model. Apart from the common involvement of humans in feature engineering, there have been also proposed some HITL techniques \cite{anderson2016runtime, gkorou2020get}. Humans can also have a role in process such as \textit{Curriculum Learning (CL)} \cite{bengio2009curriculum}, in which a scoring function organizes the dataset in terms of increasing complexity to take advantage of previously learned concepts and to ease the abstraction of new concepts. The definition of a good scoring function is one of the most complex procedures of CL, and one in which we can make use of human expertise.

\textbf{During learning} is when the HITL techniques can excel. Here we can identify three distinct categories regarding who is in control of the learning process \cite{holmberg2020feature}:
\begin{itemize}
    \item \textit{Active Learning (AL)} \cite{settles2009active}, where the system controls the learning process and relies on human input to label unlabeled data to improve model performance. 
    \item \textit{Interactive ML (IML)} \cite{amershi2014power}, characterized by closer interaction between users and learning systems, where humans frequently and interactively provide information to the system influencing in the learning process. One notable example of this type of HITL-ML because of its use in modern Large Language Models (LLMs) can be considered Reinforcement Learning from Human Feedback (RLHF) \cite{kaufmann2023survey} in which human feedback is incorporated as an integral part of the learning process, actively shaping the models’s behavior by combining environmental rewards and human insights.
    \item \textit{Machine Teaching (MT)} \cite{simard2017machine, ramos2020interactive}, where human domain experts use didactic techniques to control the learning process without needing any ML skills or expertise. This leads to curious and interesting interrelationships between human experts and ML models, which can be used as a didactic technique in itself, not to educate the model, but to educate the human itself \cite{mosqueira2023gamifying}.
\end{itemize}  
      
\textbf{After learning} there is also room for some HITL techniques. This is because nowadays is widely accepted that, if we want an AI model that to be successful, the model must also include explanatory capabilities. This is what is called \textit{Explainable Artificial Intelligence (XAI)} \cite{gunning2017explainable, abdul2018trends} and several approaches have been carried out to obtain feedback from experts with the intention of not only improving the accuracy of the results, but also the explanatory capabilities of the system \cite{guillot2022human}.

Finally, we can include techniques that are \textbf{beyond the learning process}, such as those involved in \textit{usable AI} \cite{xu2019toward} that focuses on ease of interaction with AI systems, \textit{useful AI} \cite{xu2019toward} that emphasizes practical outcomes of AI systems and \textit{trustworthy AI} \cite{choung2023trust} that ensures ethical and reliable deployment of AI systems.

\subsection{Explainable Artificial Intelligence}\label{sec:xai}

ML has already exceed human experts in specific areas, such as detecting tumors or evaluating disease progression. Although complex ML models can address complex problems, their black-box nature raises concerns about transparency and responsibility. Predictions made by these models cannot be traced back, making it unclear how or why they arrived at a certain outcome. This lack of explainability can lead to issues of trust in the ML system and the inability to provide users with human-interpretable explanations for its decisions. 

To address this issue, Explainable Artificial Intelligence (XAI) proposes a shift towards more transparent ML. Its aim is to create a set of techniques that produce more explainable models while maintaining high levels of performance in two ways: making the proper black-box model more transparent or using a transparent model to help end-users better understand model behaviors. Self-explainable modeling concerns constructing self-explanatory models at the beginning of training, e.g., building a decision tree or applying interpretability directly to the structure of the model, e.g., adding an attention layer to deep learning models.

When ML models do not allow to be self-explainable, a separate method must be conceived and applied to the model to explain its decisions. This is the purpose of post-hoc explainability techniques, which aim to communicate understandable information about how an already developed model produces its predictions for any given input. Since how a prediction is made should be as transparent as possible in a faithful and interpretable manner, model-specific and model-agnostic approaches have emerged, covering local and global interpretability \cite{barredo2020explainable}. While local explanations focus on explaining individual predictions, global explanations are the information intrinsically available in the interpretable models, such as the weights of linear models or the feature importance scores in tree models \cite{freitas2013comprehensible}.  One of the main arguments for the need for local explanations is that obtaining a global explanation of complex black-box models for all instances might be hard \cite{ribeiro2016modelagnostic}.

The problem with the post-hoc explanatory techniques is their reliability. Some authors \cite{slack2021advances} pointed out that they are inconsistent, unstable and provide very little information on their correctness and reliability, as well as being computationally inefficient.

\section{Task 1: Segmentation}\label{sec:segmentation}

\subsection{Deep Multi-Magnification Network}

The main objective of the segmentation process is to recognize the different components of the tissue slide and identify them with a color code. Table \ref{tab:color_code} shows the color code used in the segmentation process. 

\begin{table}[htbp!]
\caption{Table with the tissue components and their associated color.}
\label{tab:color_code}
\begin{tabular}{@{}cccccc@{}}
\toprule
\textbf{Component} & Stroma & Necrotic & Carcinoma & Adipose & Benign epithelial\\\hline
\textbf{Color} & Red & Blue & Yellow & Green & Orange\\
\botrule
\end{tabular}
\end{table}

The \textbf{Deep Multi-Magnification Network} model (DMMN) \cite{dmmn}, a pretrained breast cancer model that uses partial annotations as inputs, was used to automatically segment the different tissue components by using a set of patches from multiple magnifications of the whole slide images. The steps that define this model are the following:

\begin{itemize}
    \item \textbf{Partial annotations}. The objective is to annotate some regions of the WSI and identify the different components of the tissue. The annotation of the different regions of a WSI is done by an expert. As this operation requires a lot of time and effort, to significantly reduce them, the partial annotations done by the expert are going to be used to train a model in further steps. The different tissue components and the color associate to them are shown in Table \ref{tab:color_code}.
    \item \textbf{Training patch extraction}. WSIs are generally too large to be processed by a CNN. To analyze them, patch-based methods where used, dividing the WSI in smaller patches and processing them with a CNN, combining the output of each patch to form an image.
    \item \textbf{Class balancing}. When the number of patches of one tissue component is significantly higher than another tissue component, the CNN can not correctly learn the minor one. To solve this problem, a class balancing technique is applied, the elastic deformation technique \cite{elastic_deformation}, that creates new images applying local deformations to the original images. This technique is applied to each patch.
    \item \textbf{Training}. The CNN is trained with all the images obtained in the previous steps.
    An example of the output of the trained model is shown in Figure \ref{fig:output_example_segmentation}.
    
\begin{figure*}[htbp!]
    \centerline{\includegraphics[width=0.9\textwidth]{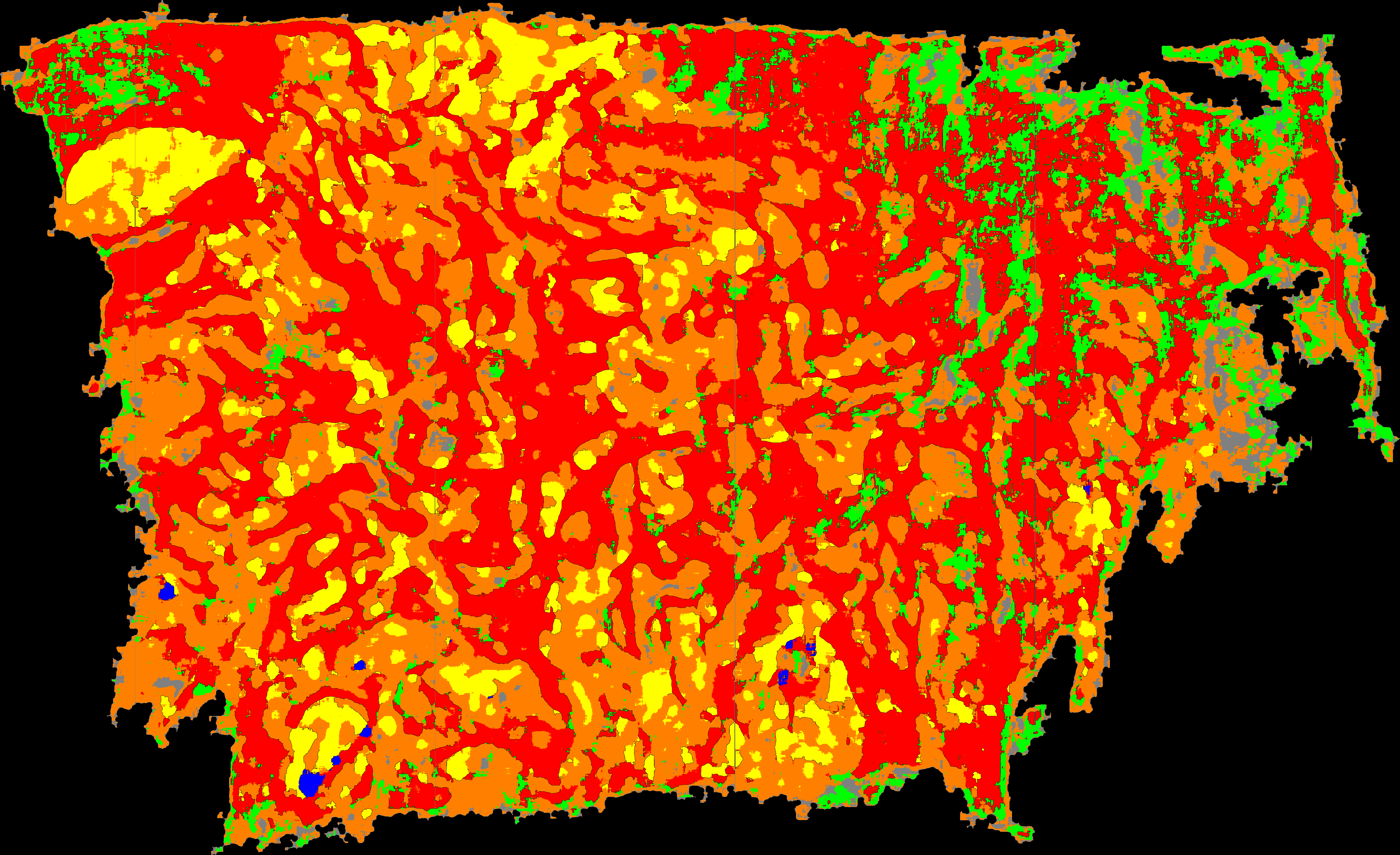}}
    \caption{An example of the output of the segmentation model.}
    \label{fig:output_example_segmentation}
\end{figure*}

\end{itemize}

\subsection{HITL for segmentation correction}

The result of the segmentation process was evaluated by a human expert, a pathologist, who was responsible for identifying erroneous segmentations and correcting them. A total of 19 images were analyzed and corrected by the pathologist expert, conformed by over 3 images of each cancer type. Figure \ref{fig:example_images_expert} shows an example of the images that were provided to the pathologist to evaluate, where:

\begin{figure}[htbp!]
    \centering
    \begin{subfigure}{\textwidth}
        \includegraphics[width=\textwidth]{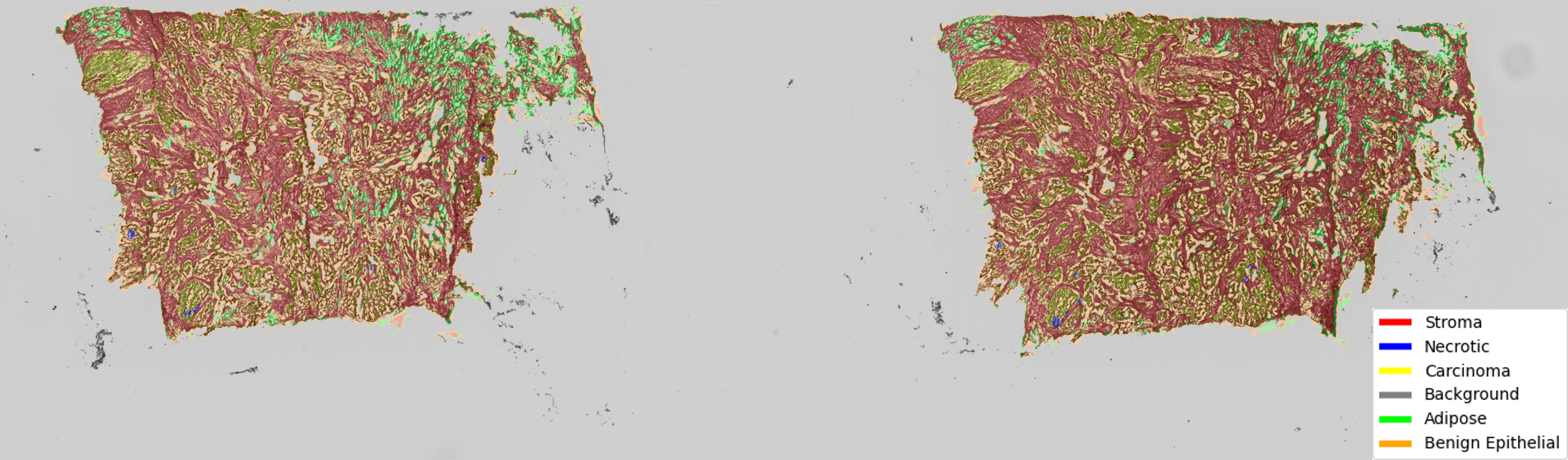}
        \caption{}
        \label{}
    \end{subfigure}
    \hfill
    \begin{subfigure}{\textwidth}
        \includegraphics[width=\textwidth]{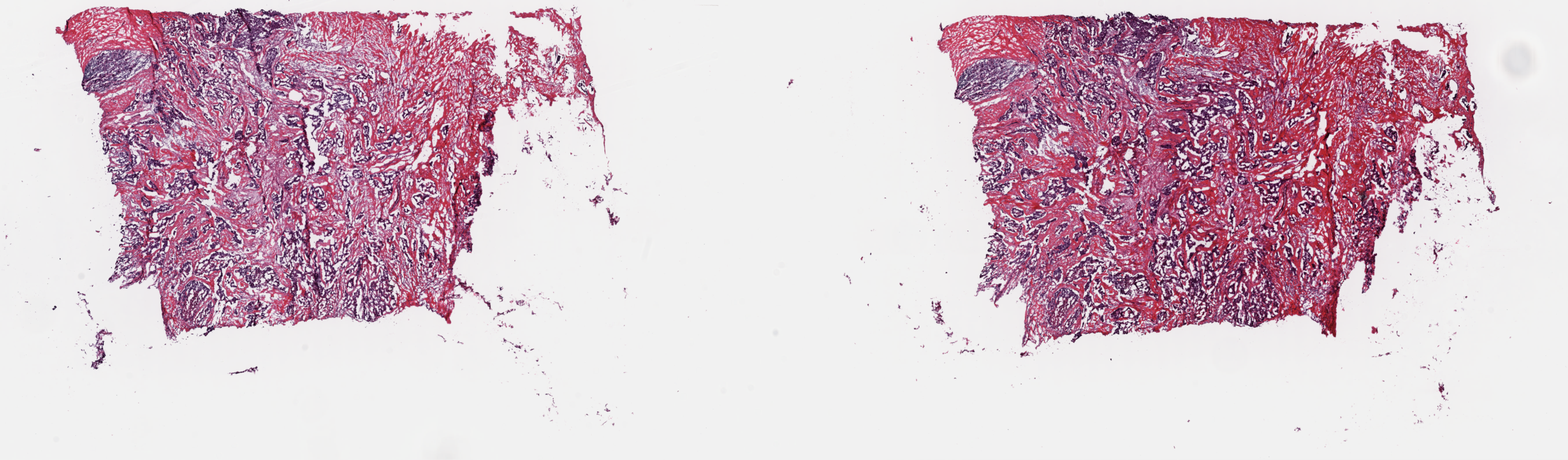}
        \caption{}
        \label{}
    \end{subfigure}
    \caption{(a) Segmented image obtained by the network superposed over the original one and (b) original Whole Slide Image (WSI).}
    \label{fig:example_images_expert}
\end{figure}

\begin{itemize}
    \item the WSI is the image that experts usually use to identify cancer nodes and it was requested by the pathologist.
    \item the result of the segmentation superimposed on the original image to facilitate the pathologist's task.
\end{itemize}

\begin{figure}[htbp!]
\centering
\includegraphics[width=\textwidth]{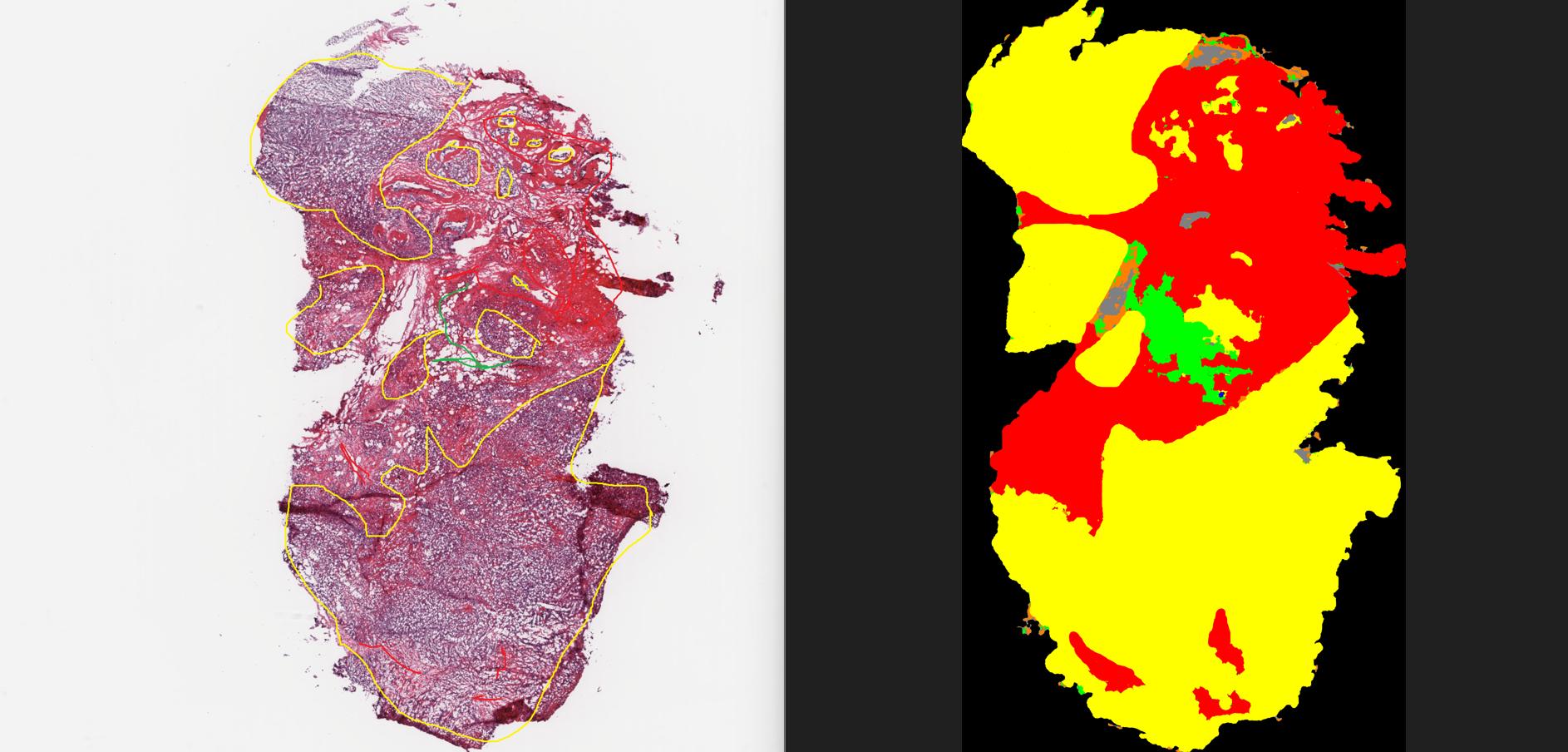}
\caption{Corrections proposed by the pathologist (left) and segmented image after applying the pathologist's corrections.}
\label{fig:segmented_after_hitl}
\end{figure}

A total of 19 images underwent analysis by the pathologist. Figure \ref{fig:segmented_after_hitl} illustrates an example with the proposed corrections, along with the corrected image that followed. The pathologist meticulously reviewed the 19 segmented images, annotating regions on the whole-slide image (WSI) where he deemed necessary. By annotating the segmented image, a new version was generated, adhering to the corrections suggested by the pathologist. These adjustments effectively simplified the information within the image, compacting various areas while preserving essential high-level features.

The analysis process carried out by the pathologist to correct the segmentation errors led to the following conclusions:
\begin{itemize}
    \item The segmentation images analyzed showed a maximum accuracy in identifying tumor nests, which were the part that the pathologist was most interested in.
    \item The segmentation images analyzed showed a non-maximum accuracy in identifying desmoplastic/neoplastic stroma and necrotic areas.
    \item The main deficit is recognizing the normal adipose tissue and normal ductal/glandular structures.
\end{itemize}

\section{Task 2: Classification}\label{sec:classification}

\subsection{Models}

The selection of a model is a key aspect of the classification task. In this case, we have decided to use pretrained models and transfer learning thanks to their accessibility and the resource efficiency, avoiding the need to train a new model from scratch, but mainly because their high capability to extract patterns from any kind of image. 

\begin{figure}[htbp!]
\centering
\includegraphics[width=\textwidth]{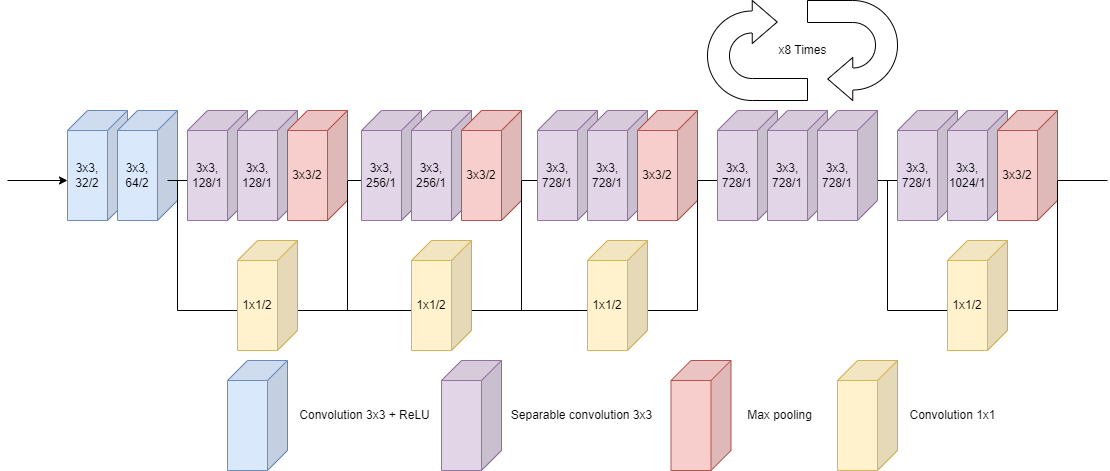}
\caption{Architecture of the Xception model. The numbers in each block represent the kernel size, the number of filters in the current block and the stride.}
\label{fig:xception_arch}
\end{figure}

The first model that was selected is Xception, a CNN architecture firstly introduced in 2016 by François Chollet \cite{DBLP:journals/corr/Chollet16a}. The name Xception is derived from ``Extreme Inception'', since it uses the Inception architecture principles and adds several innovations, such as depthwise separable convolutions, depthwise convolutions and pointwise convolutions, reducing significantly the number of parameters of the network while having similar computational capabilities to regular convolutions. Figure \ref{fig:xception_arch} shows the architecture of the Xception model.

\begin{figure}[htbp!]
\centering
\includegraphics[width=\textwidth]{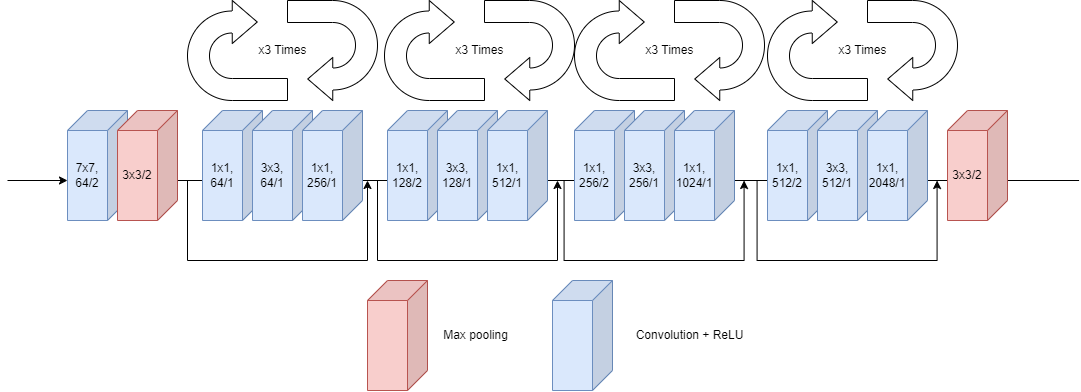}
\caption{Architecture of the Resnet model. The numbers in each block represent the kernel size, the number of filters in the current block and the stride.}
\label{fig:resnet_arch}
\end{figure}

The other model is Resnet50, one of the most widely used CNN architectures that was firstly introduced in 2015 by Kaiming He et al. \cite{DBLP:journals/corr/HeZRS15}. The Resnet networks are formed by multiple residual blocks, that helps avoiding the vanishing gradient problem passing the input of some layers to deeper layers. This version of Resnet has a total of 50 layers, having a considerable capacity solving computer vision problems. Figure \ref{fig:resnet_arch} shows the architecture of the Resnet50 model.

The metrics used to analyze the results of the process are Accuracy and F1-score. The first one is the most common metric and will analyze the proportion of correctly classified instances out of the total of instances, providing a general overview of the performance of the trained model across all classes. The second one is really useful for imbalanced datasets, giving a more robust measure of the model's performance in those scenarios. 

\subsection{Experiments}
Using the previous explained models and the images downloaded a total of 4 experiments were done. The main difference between the 4 experiments is whether they use the segmented images or if they add the images analyzed by the pathologist and the number of cancer types to classify.

\subsubsection{Experiments with the segmented images}

The first two experiments used the images obtained directly from the segmentation model. Although the experiments use different amount of images, the train, validation and test subsets are composed by the 60\%, 20\% and 20\% of the data in both cases, respectively. The pretrained model used in both cases is the Xception one with the \textit{ImageNet} weights. The last two convolutional blocks of the pretrained model will be retrained, maintaining the \textit{ImageNet} weights in the rest of them. After the pretrained model layers, a flatten layer followed by 2 dense layers with the \textit{ReLU} activation function, a dropout layer for regularization purposes and 1 dense layer with the \textit{softmax} activation function were added. The added layers will process the extracted features obtained from the pretrained model to classify them one of the target classes. Table \ref{tab:hyperparameters_exp1} shows the hyperparameters that were selected for these experiment.

\begin{table}[htbp!]
\caption{Table with the hyperparameters of experiments 1 and 2.}
\label{tab:hyperparameters_exp1}
\begin{tabular}{@{}ccccc@{}}
\toprule
\textbf{Batch size} & \textbf{Dropout rate} & \textbf{Learning rate} & \textbf{Input shape} & \textbf{Optimizer}\\\hline
32 &  0.5 &  0.0001 & (224, 224) & Adam \\
\botrule
\end{tabular}
\end{table}

The specific characteristics of each experiment are explained in the following paragraphs:

\paragraph{Experiment 1} 

Around 900 images were used in this experiment. The mentioned images can belong to one of the following 4 cancer types: Basal, Her2, Luminal A and Luminal B. 

\paragraph{Experiment 2} 

Around 780 images were used in this experiment. The mentioned images can belong to one of the following 3 cancer types: Basal, Luminal A and Luminal B. The previous experiment was unable to correctly classify any of the images belonging to the Her2 class. On top of that, it is the class with the less amount of images and one of the most difficult to classify. For those reasons, the Her2 class was excluded from this experiment.

\subsubsection{Experiments with the images analyzed by the pathologist}

These experiments use the images obtained by the segmentation model, but they also introduce the segmentation images corrected by the pathologist to the training set. Our objective was to see if the corrected segmentation images could help with the learning of key features and the generalization of the model during the training process. Although the experiments use different amount of images, the train, validation and test subsets are composed by the 60\%, 20\% and 20\% of the data in both cases, respectively. 
The specific characteristics of each experiment are explained in the following paragraphs:

\paragraph{Experiment 3} 

Around 900 images were used in this experiment. The mentioned images can belong to one of the following 4 cancer types: Basal, Her2, Luminal A and Luminal B. The pretrained model used in this case is the Xception one with the \textit{ImageNet} weights. The architecture used in this experiment is the same that was used in the previous 2 experiments.  Table \ref{tab:hyperparameters_exp3} shows the hyperparameters that were selected for this experiment.

\begin{table}[htbp!]
\caption{Table with the hyperparameters of experiment 3.}
\label{tab:hyperparameters_exp3}
\begin{tabular}{@{}ccccc@{}}
\toprule
\textbf{Batch size} & \textbf{Dropout rate} & \textbf{Learning rate} & \textbf{Input shape} & \textbf{Optimizer}\\\hline
32 &  0.404 &  0.0001 & (224, 224) & Adam \\
\botrule
\end{tabular}
\end{table}

\paragraph{Experiment 4} 

Around 780 images were used in this experiment. The mentioned images can belong to one of the following 3 cancer types: Basal, Luminal A and Luminal B. In order to compare this experiment with experiment 2, the class Her2 was discarded. The pretrained model used in this case is the Resnet50 one with the \textit{ImageNet} weights. After the pretrained model layers, a flatten layer followed by 3 dense layers with the \textit{ReLU} activation function, a dropout layer and, finally, a dense layer with the \textit{softmax} activation function were added. Table \ref{tab:hyperparameters_exp4} shows the hyperparameters that were selected for this experiment.

\begin{table}[htbp!]
\caption{Table with the hyperparameters of experiment 4.}
\label{tab:hyperparameters_exp4}
\begin{tabular}{@{}ccccc@{}}
\toprule
\textbf{Batch size} & \textbf{Dropout rate} & \textbf{Learning rate} & \textbf{Input shape} & \textbf{Optimizer}\\\hline
46 &  0.404 &  0.0001 & (256, 256) & Adam\\
\botrule
\end{tabular}
\end{table}

\subsection{Results}

This section will present and discuss the results obtained by each of the previously explained experiments. Table \ref{tab:results_classification} shows the accuracy and the F1-score obtained by each experiment.

Overall, the results obtained were not good, showing that the model struggles differentiating between the different classes, even adding segmentation images corrected by a pathologist to the training process. The classes that were more difficult to classify were the Her2 and Basal ones, probably because the amount of images belonging to these cancer types were limited compared with the Luminal A and Luminal B cancer types. 

The corrected segmentations showed mixed performances, helping in the experiments that used 4 cancer types, but penalizing the experiments with 3 cancer types.

\begin{table}[htbp!]
\caption{Table with accuracy and F1-score metrics of each experiment}
\label{tab:results_classification}
\begin{tabular}{@{}ccc@{}}
\toprule
\textbf{Experiment} & \textbf{Accuracy} & \textbf{F1-score} \\\hline
  Experiment 1 & 0.34 &  0.24 \\
  Experiment 2 & 0.43 &  0.36 \\
  Experiment 3 & 0.39 &  0.18 \\
  Experiment 4 & 0.33 &  0.20 \\
\botrule
\end{tabular}
\end{table}

Cancer classification directly from images is a complex task that is still under research. For pathologists, the detection of cancer regions directly from images is quite difficult and time consuming. The state of art for breast cancer detection is to analyze 50 well-known genes and determine the cancer type that the patient has.
For the previous reasons, it is normal that the results obtained were suboptimal.

\section{Task 3: Interpretation}\label{sec:interpretation}

\subsection{LIME}\label{sec:xai_lime}

LIME or Local Interpretable Model-agnostic Explanations \cite{lime2016why} is a novel explanation technique that explains predictions of any classifier in an interpretable and faithful manner by learning an interpretable model locally around the prediction.

Intuitively, an explanation is a local linear approximation of the model's behavior. While the model may be very complex globally, it is easier to approximate it around the vicinity of a particular instance. While treating the model as a black box, the instance to explain is perturbed and a sparse linear model around it is learned as an explanation. The operation of the method is as follows:
\begin{itemize}
\item \textbf{Generate a perturbed dataset}: Each training instance is modified and a new dataset with the modified instances is created. For image instances, different regions of the image are modified hiding their pixels. 
\item \textbf{Classify the perturbed dataset}: the original ML model is used to classify the instances of the perturbed dataset. 
\item \textbf{Calculate distances}: The distance between the original instance and the perturbed instance is calculated using cosine similarity and then normalized.
\item \textbf{Train of a linear explainable model}: Using the perturbed dataset, the original ML model predictions and the distances previously calculated a linear regression model is trained. The most likely prediction will be the class to which the LIME explanation corresponds. The coefficients of the resulting linear regression model with the largest magnitude are those that have the greatest significance in the prediction of the model.
\end{itemize}

A weak point of this approach is the required transformation of any type of data in a binary format that is claimed to be human interpretable. Moreover, in practice the explanation is only provided through linear models and their features importance \cite{guidotti2019asurvey}.

\subsection{SHAP}\label{sec:xai_shap}

SHAP (SHapley Additive exPlanations) \cite{lundberg2017aunified} is a method to explain individual predictions, based on the game theoretically optimal Shapley values. Shapley values \cite{watson2023explaining} are a widely used approach from cooperative game theory that come with desirable properties. The feature values of a data instance act as players in a coalition. The Shapley value is the average marginal contribution of a feature value across all possible coalitions.

SHAP creates explanations of the model by asking for each prediction and feature the following question: ``How does prediction $x$ change when feature $y$ is removed from the model?'' The answer is Shapley values, which quantify the magnitude and direction (positive or negative) of the effect of a feature on a prediction. Shapley values can produce model explanations with the clarity of a linear model.

When calculating Shapley values, we calculate the contribution of a feature by determining its marginal contributions to all possible subsets and then taking the weighted average. While this may be feasible for models with a small number of input features, as the number of input features increases, the cost of this calculation is sufficiently high and, in some cases, impossible to obtain. This is why Shapley values are usually approximated by the closest values using KernelSHAP \cite{lundberg2017aunified}.

KernelSHAP is a method that uses the LIME technique to calculate the Shapley values.  With KernelSHAP, a sufficient number of coalition vectors (simplified features) is created. In each vector, features with a value of 1 are replaced by their actual values, and features with a value of 0 are replaced by different feature values. For this instance, the prediction of the ML model is obtained and the weight of each feature is calculated. A linear model is trained using these weight values whose target value is the predicted value of the ML  model for that sample. The coefficient values of the linear model correspond to Shapley Values for each feature.

SHAP offers both global and local interpretability. It provides a holistic understanding of the overall model behavior by analyzing feature importance globally. Additionally, it allows for specific explanations on individual predictions, enhancing fine-grained interpretability.
 
\subsection{Grad-CAM}\label{sec:xai_grad_cam}

Gradient-weighted Class Activation Mapping (Grad-CAM) \cite{selvaraju2017gradcam}, is a technique for producing \emph{visual explanations} for decisions from a large class of Convolutional Neural Network (CNN)-based models, making them more transparent. 

The Class Activation Mapping (CAM) \cite{zhou2016learning} method identifies discriminative regions used by a restricted class of image classification CNNs which do not contain any fully-connected layers. This approach is a generalization of CAM and is applicable to a significantly broader range of CNN model families.

The Grad-CAM uses the gradients of any target concept, flowing into the final convolutional layer to produce a coarse localization map highlighting the important regions in the image for predicting the concept by means of a heat map. For this purpose, the algorithm analyses the last convolution layer of the model, which is the one that captures the high-level features and contains information about the important regions of the input image. The outputs of this last convolution layer, which are several feature maps, are computed together with the prediction, to result in as many gradients as feature maps.

Once these gradients are obtained, the next step is to compare the different feature maps in terms of relevance using the Global Average Pooling \cite{lin2014network}. The higher the average on a given feature, the greater the influence of that image feature on the prediction. Once we have the rating of each feature map, we proceed to obtain the heat map. 
To do this, the weighted sum of the feature maps is performed and used as input to a ReLU function (this activation function is used since only the values with a positive influence on the prediction are of interest) which returns the heat map. 
The final step involves superimposing the heat map on the original image to identify the areas of the image most relevant to the prediction.

Extensive human studies reveal that the Grad-CAM visualizations can discriminate between classes more accurately, better reveal the trustworthiness of a classifier, and help identify biases in datasets \cite{selvaraju2017gradcam}.

\subsection{HITL optimization}\label{sec:xai_hitl}

\subsubsection{Bayesian hyperparameter optimization}

As we know, the hyperparameters of a ML model are those variables whose values are set prior to the application of the model on the data and whose values are not adjusted during the training of the model \cite{bengio2012practical}. They are parameters that can be adjusted manually or by algorithms and that control the behavior of the model and/or determine its structure.

There are many hyperparameters in a ML model, and finding the optimal value for them is a complex task that is usually tried to be automated by means of optimization mechanisms. In our particular case we decided to include the following hyperparameters in the optimization process.

\begin{itemize}
    \item \textbf{Dropout}: It specifies the percentage of neurons to be deactivated in a layer during a training iteration. It is used to avoid network overfitting, which occurs when the model fits too closely to the training data and its noise, thus losing generalization power. A small dropout rate may prevent less overfitting, while a large dropout rate may produce models with lower pattern learning capacity.
    \item \textbf{Learning rate}: Determines how much the model weights should be adjusted as a function of the gradient of the loss function during each iteration of training. A small learning rate slows down the training but smoothes the convergence, while a high learning rate speeds up the learning but convergence may not be achieved.
    \item \textbf{Batch size}: Determines the number of samples used in each iteration of the training. A small batch size may produce greater variance between iterations as the weights between one set of samples and another may be very different when updated. A large batch size requires more memory and may be less efficient in small datasets.
\end{itemize}

The goal of hyperparameter optimization (also known as hyperparameter tuning) is to find the optimal set of hyperparameters for a given learning algorithm  \cite{feurer2019hyperparameter}. The simplest way to perform this task is to carry out a search. It could be a \textit{Grid Search}, which involves exhaustively exploring a manually specified subset of the hyperparameter space; or a \textit{Random Search}, that instead of exhaustive enumeration, selects hyperparameter combinations randomly.

A more advance method for hyperparameter optimization is Bayesian Optimization (BO) \cite{bergstra2011algorithms}. This method consists of considering an objective function that we want to optimize that it is typically a performance metric (accuracy in our case) of the ML model given some hyperparameters. If we were to know the probability distribution of this objective function, then we can simply compute the gradient descent and find the global minimum or maximum (depending on the metric used). However, since we don't know the shape of this function we need to use techniques, such as BO. 

BO is a probabilistic optimization technique that aims to find the global minimum (or maximum) of an objective function using a surrogate model (often by means of a Gaussian Process model \cite{bishop2006pattern}) to approximate the true objective function. The surrogate model guides the search by suggesting where to evaluate the objective function next (applying an Expected Improvement process \cite{chen2023hierarchical}). As more evaluations are done, the surrogate model gets refined, leading to better decisions. This method is considered appropriate for tuning hyperparameters because it can find and optimal value of a surrogate function with only a small number of samples \cite{wu2019hyperparameter}.

For the BO process to work correctly it is necessary to establish the range of values of each hyperparameter with which the Bayesian optimization will test different configurations. These ranges are quite wide since in the first optimization we seek to achieve the configuration that maximizes the accuracy at its highest possible value without taking so much into account the execution time. The ranges chosen for our hyperparameters were:

\begin{itemize}
    \item \textbf{Dropout}: [0.05, 0.5]
    \item \textbf{Learning rate}: [0.00001, 0.1]
    \item \textbf{Batch size}: [1, 32]
\end{itemize}

With the objective function and the hyperparameter ranges defined, the last step is to perform the Bayesian optimization itself. In our case it was done using the Bayesian Optimization package \cite{nogueira2014bayesian}. 

\subsubsection{Human-in-the-Loop hyperparameter tunning}

Bayesian optimization has been successful applied in several areas, apart from hyperparameter tuning, such as A/B Testing, recommender systems, robotics and reinforcement learning, etc. \cite{shahriari2016taking}. 
Some of these applications involves the use of human feedback to drive the process. For example, in \cite{brochu2010bayesian}, the authors use Bayesian optimization to set the parameters of several animation systems by showing the user examples of different parameterized animations and asking for feedback. This interactive Bayesian optimization strategy is particularly effective as humans can be very good at comparing examples, but unable to produce an objective function whose optimum is the example of interest.

Another work \cite{kim2017human} used human-in-the-loop Bayesian optimization for quickly identifying near-optimal control parameters based on real-time physiological measurements. The experiment consisted of minimizing metabolic cost by optimizing walking step frequencies in unaided human subjects and it proved that Bayesian optimization could be used in the context of Human-Computer Interaction (HCI) to improve system convergence and lower overall energy expenditure.

Finally, in \cite{guillot2022human} a human-in-the-loop optimization technique was carried out not only to improve the performance of the model but also render its decision-making more interpretable and explainable to human users. In our case we followed a similar approach and we assessed the quality of the interpretability explanations returned by the XAI algorithms using domain experts (in this case, a pathologist). It is this pathologist's assessment that will guide the hyperparameter optimization process by Bayesian optimization. The process is composed of the following phases:

\begin{enumerate}
    \item \textbf{BO of hyperparameters of the classification model}: An initial optimization is performed following a Bayesian optimization approach and we are left with the models with the highest accuracy in the validation set.
    \item \textbf{Application of XAI methods}: The selected XAI methods are applied to each of the models to obtain an interpretation of what part of the original figure was used by the model to obtain a final classification.
    \item \textbf{Selection of the set of images to be evaluated}: An image set consisting of the original biopsy image, its segmentation and its interpretation by each of the models is selected.
    \item \textbf{Evaluation of the pathologist interpretation}: For each of the images in the previous set, the pathologist will evaluate each XAI method interpretation with a grade based on his own rubric and shown in Table \ref{tab:rubric}, where 5 is the best punctuation and 0 the minimum one. 
    \item \textbf{Analysis of results}: Once the pathologist scores each image interpretation, we select the model with the best interpretability and explainability, taking into account other metrics such as accuracy or loss in case of a tie. 
\end{enumerate}

\begin{table}[htbp!]
\caption{Table with the evaluation rubric described by the pathologist}
\label{tab:rubric}
\begin{tabular}{@{}cp{0.8\linewidth}@{}}
\toprule
    \textbf{Grade} & \textbf{Criteria} \\\hline
    \textit{0} & The area of interest, highlighted in green, does not appear or is completely outside the biopsy. \\
    \textit{1} & The area of interest is almost entirely outside of the biopsy. \\
    \textit{2} & The area of interest is both inside and outside the biopsy, but without matching to the shape of the tissue. \\
    \textit{3} & The area of interest is mostly within the biopsy and matches the shape of tissues of little relevance to tumor staging \\
    \textit{4} & The area of interest is mostly within the biopsy and matches the shape of relevant tissues, although also some less relevant ones. \\
    \textit{5} & The area of interest is completely within the biopsy and fits perfectly with relevant areas such as the carcinoma. \\
\botrule
\end{tabular}
\end{table}

This process is iterative and is repeated each time, making the variation margins of the selected hyperparameters smaller and smaller, so that each time we get closer to their optimal value. Since the classification results of the different models were far from optimal we decided to focus the objective of the optimization process in augmenting the quality of the interpretation. But the same technique can be used for improving not only interpretation but also classification metrics.

\subsection{Results}\label{sec:xai_results}

In this section we will show the interpretability results obtained with the different explainability models. We will show some examples of the optimization process and the final results obtained.

\subsubsection{Selection of the explainability method}

\begin{figure*}[htbp!]
    \centerline{\includegraphics[width=\textwidth]{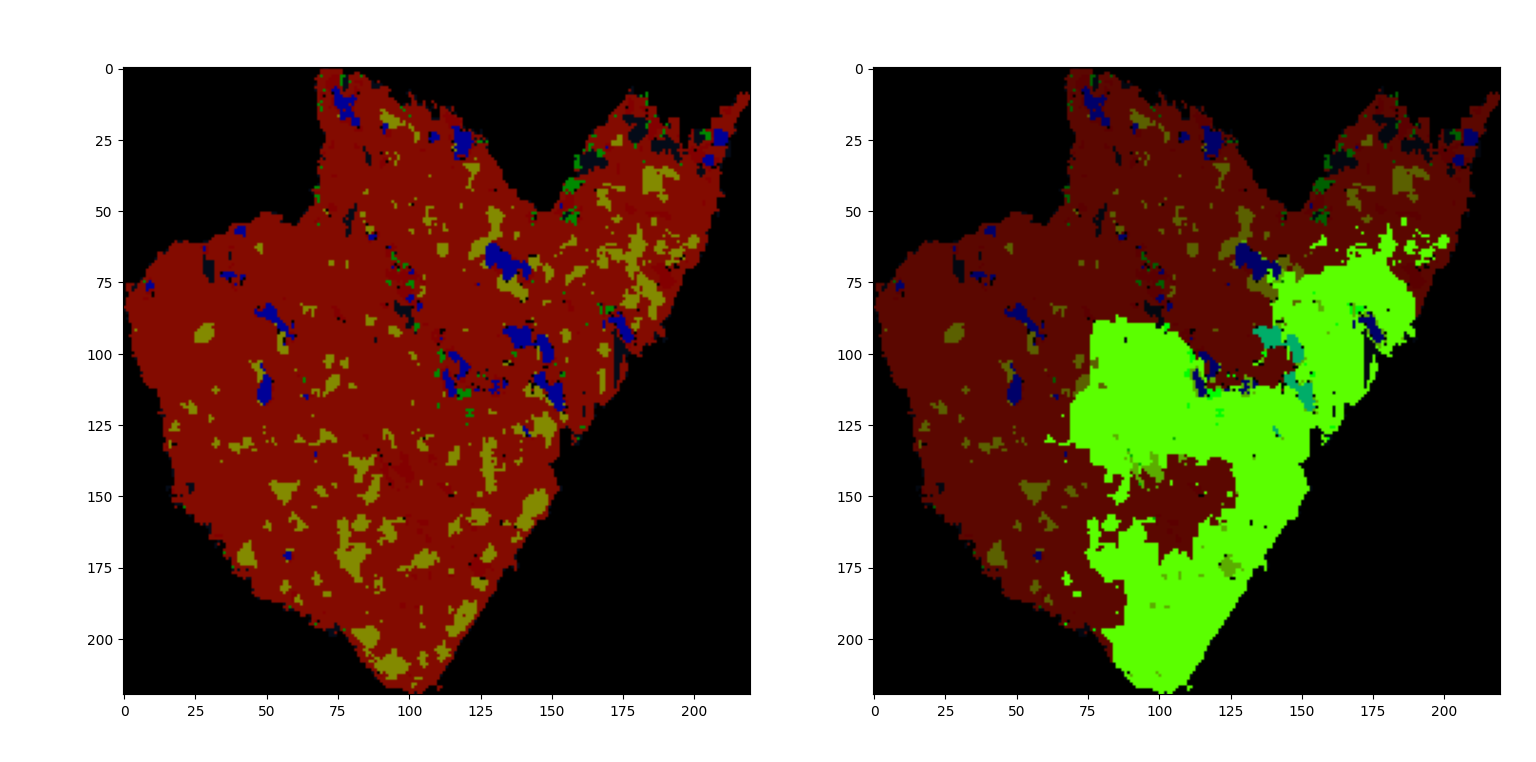}}
    \caption{LIME results for a given example.}
    \label{fig:lime_example}
\end{figure*}

The first step was to compare the results of the different explainability methods and draw some conclusions. The \textbf{LIME} method returns interpretations such as those in Figure \ref{fig:lime_example}. In it we can see the area of interest of the image, in green color, where (according to LIME) the classification model is being focused to classify it in this case as Basal type. It is an interpretation that allows the expert, observing several images belonging to different classifications, to draw conclusions of how the model is behaving internally.

\begin{figure*}[htbp!]
    \centerline{\includegraphics[width=\textwidth]{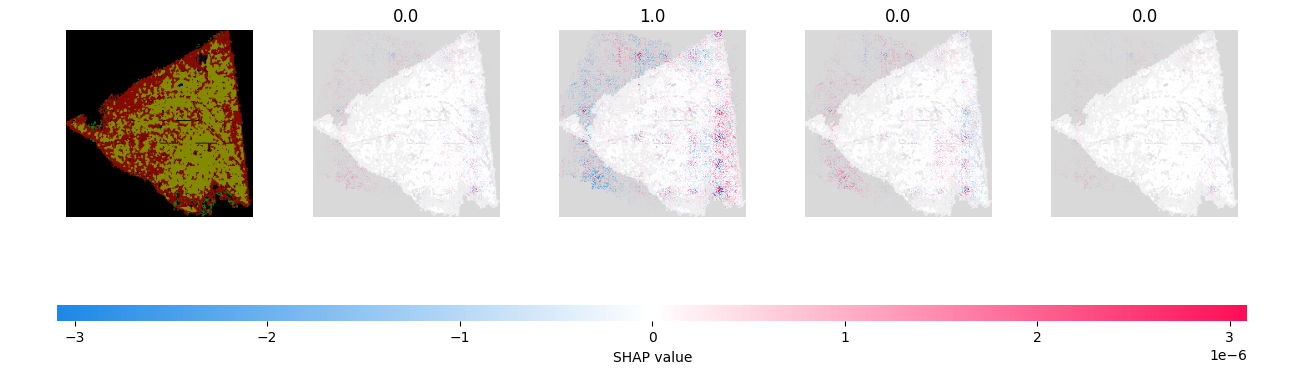}}
    \caption{SHAP results for a given example.}
    \label{fig:shap_example}
\end{figure*}

The \textbf{SHAP} method obtains results such as those shown in Figure \ref{fig:shap_example}. It can be seen that both the regions of interest that determine the classification of the model (in this case, in red) and those that do not reflect the classification provided by the model (in blue) are scattered over the image without discriminating between biopsy area or image background. Moreover, these regions of interest are scarce. These results are very inadequate, so it was decided not to rely on SHAP interpretability in the optimization process with the collaboration of human experts.

\begin{figure*}[htbp!]
    \centerline{\includegraphics[width=0.5\textwidth]{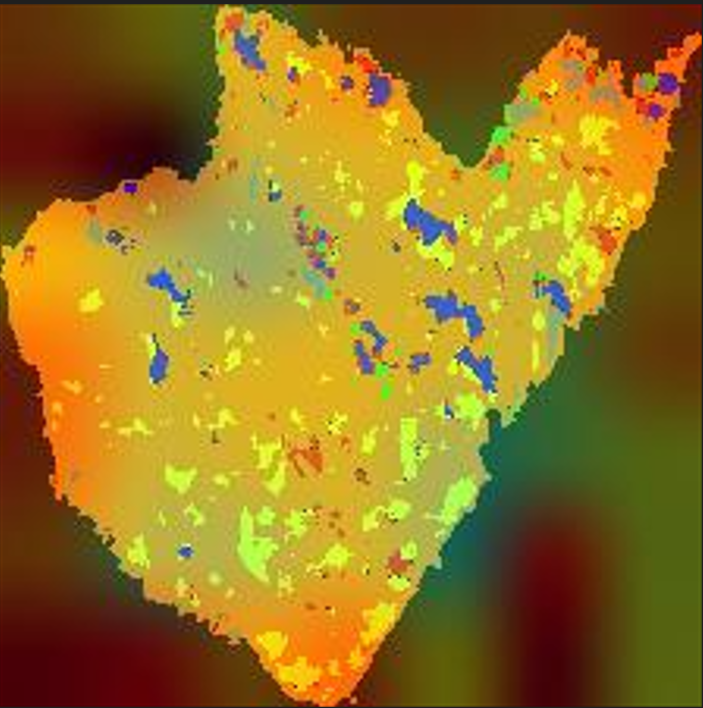}}
    \caption{Grad-CAM results for a given example.}
    \label{fig:gradcam_example}
\end{figure*}

The \textbf{Grad-CAM} method was also discarded since it was not possible to obtain accurate interpretations that would help to understand the model. Consistent results were not even obtained for different predictions on the same image. In Figure \ref{fig:gradcam_example} we can see that the heat map superimposed on the original image has several issues: it highlights areas outside the image and inside the image the highlights do not conform with the tissue structures present in the biopsy. Here warm colors such as the red and orange correspond to a high score for a given class, while cold colors, such as green and blue correspond to small evidence for the class. 

These problems with the interpretability algorithms may be also a consequence of the problems that the ML models have to correctly classify each cancer type. But, since the LIME results were the only ones with enough quality to be presented to the pathologist, these were the results that were used in the Bayesian optimization of hyperparameters. 

\subsubsection{Optimization of the explainability model}

The hyperparameter optimization process started in the first iterations without any human intervention in order to discard those combinations that offer worse results. The results obtained can be seen in Table \ref{tab:config_param} from which we decided to discard models 3 and 4 and focus the expert feedback on models 1, 2 and 5.

\begin{table}[htbp!]
\caption{Accuracy results of the different models considered with their corresponding hyperparameters}
\label{tab:config_param}
\begin{tabular}{@{}ccccc@{}}
\toprule
\textbf{Model} & \textbf{Batch size} & \textbf{Dropout rate} & \textbf{Learning rate} & \textbf{Accuracy}\\\hline
  Model 1 &  6 &  0.404 &  0.00175 &  0.38 \\
  Model 2 &  4 &  0.405 &  0.00100 &  0.38 \\
  Model 3 &  1 &  0.126 &  0.00878 &  0.27 \\
  Model 4 &  1 &  0.123 &  0.00888 &  0.26 \\
  Model 5 &  7 &  0.425 &  0.00019 &  0.38 \\
\botrule
\end{tabular}
\end{table}

In every human-in-the-loop approach there is a trade-off between the amount of cases an expert can collaborate on (the more the better), and the amount of time and effort the expert will invest in analyzing those cases (the less the better). Therefore, we want to draw the attention of the expert to those more promising models, and we want to select those cases that are representative of the whole problem in hand. 

Thus, finally we narrowed down our problem to optimizing three models based on LIME interpretability results. We give to the pathologist three images for each cancer type and for each ML model and we ask him to rate their LIME results (using the rubric of Table \ref{tab:rubric}).

\begin{figure}[htbp!]
    \centering
    \begin{subfigure}{0.4\textwidth}
        \includegraphics[width=\textwidth]{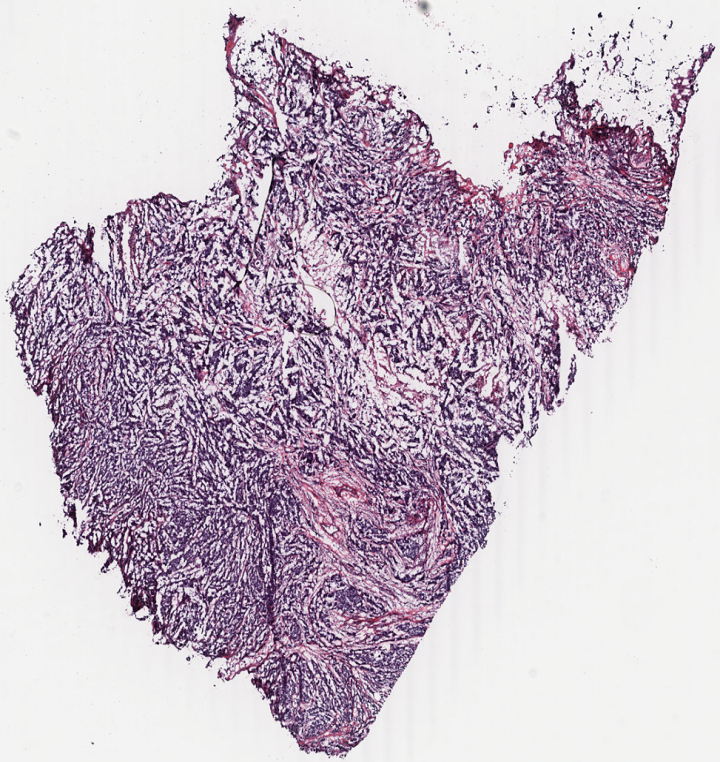}
        \caption{}
        \label{fig:image01}
    \end{subfigure}
    \hfill
    \begin{subfigure}{0.44\textwidth}
        \includegraphics[width=\textwidth]{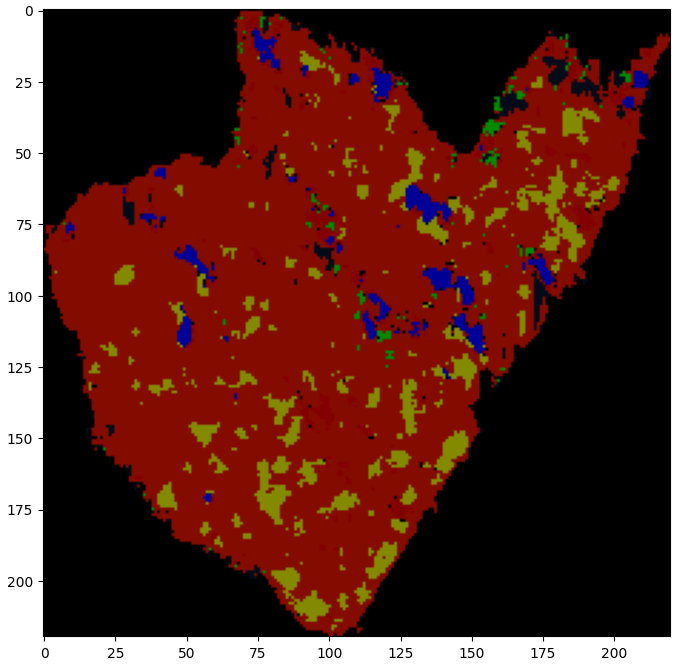}
        \caption{}
        \label{fig:image01_segmented}
    \end{subfigure}
    \caption{(a) Biopsy and (b) segmentation of a basal type image}
    \label{fig:image01_original_segmented}
\end{figure}

As an example of the work carried out by the pathologist we can see in Figure \ref{fig:image01_original_segmented} the original image of a biopsy and the segmented version of it. With this information, the pathologist was asked to rate the explainability results of three models (1, 2 and 5) each of one presented the same overall accuracy.

\begin{figure}[htbp!]
    \centering
    \begin{subfigure}{0.32\textwidth}
        \includegraphics[width=\textwidth]{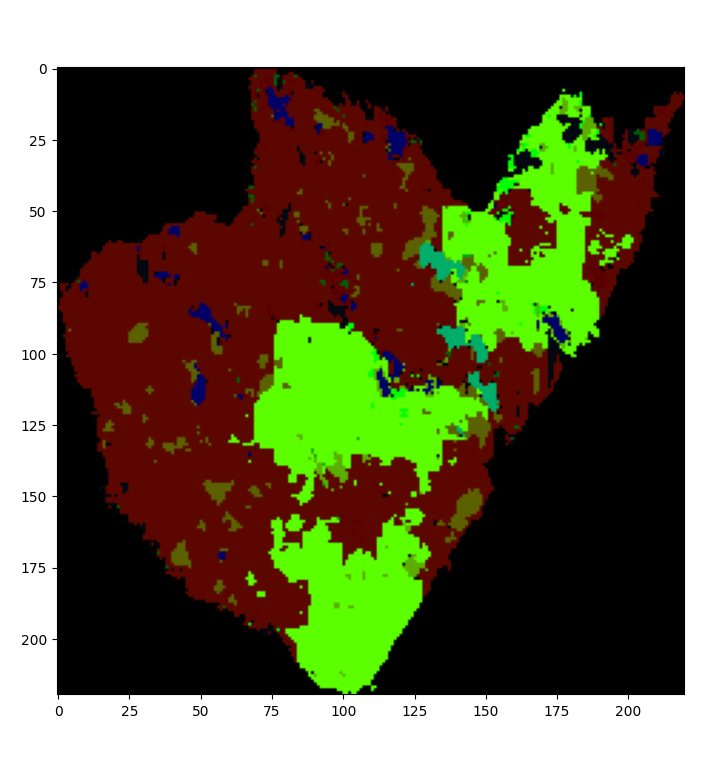}
        \caption{Model 1}
        \label{fig:image01_model1}
    \end{subfigure}
    \hfill
    \begin{subfigure}{0.32\textwidth}
        \includegraphics[width=\textwidth]{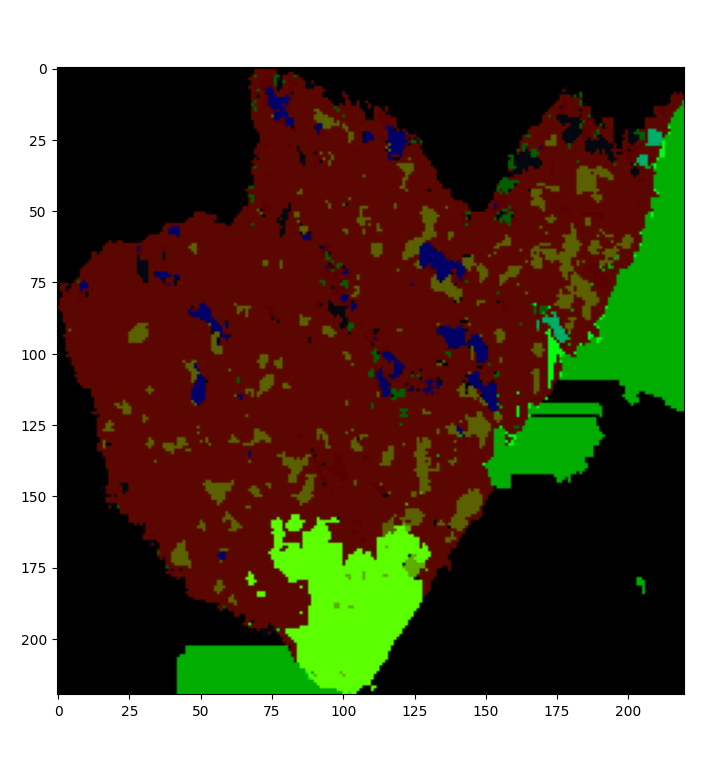}
        \caption{Model 2}
        \label{fig:image01_model2}
    \end{subfigure}
    \hfill    
    \begin{subfigure}{0.32\textwidth}
        \includegraphics[width=\textwidth]{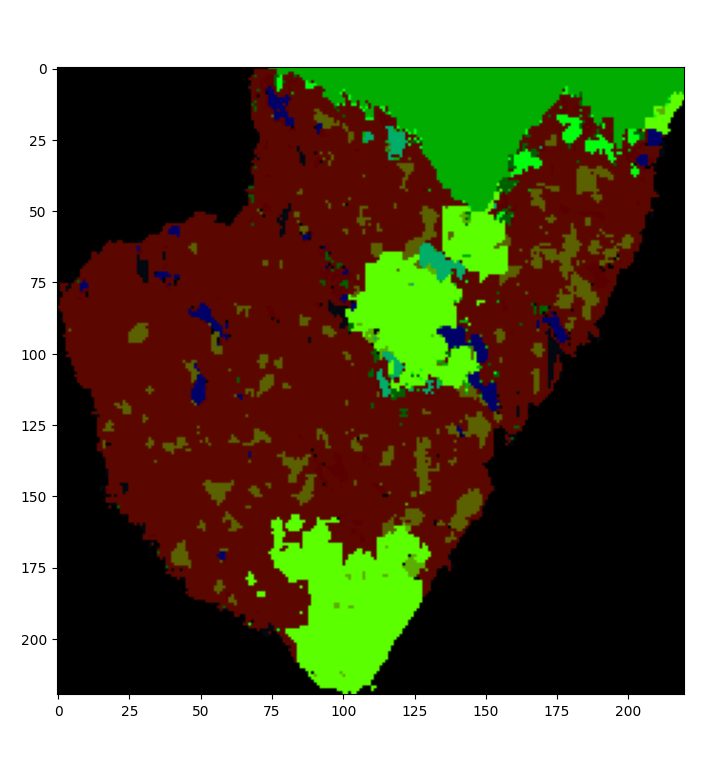}
        \caption{Model 5}
        \label{fig:image01_model5}
    \end{subfigure}    
    \caption{Explainability results of different models for the same image}
    \label{fig:imagen01_LIME_results}
\end{figure}

The LIME explainability results of the three models can be seen in Figure \ref{fig:imagen01_LIME_results} where we can easily see that the best model in terms of explainability is Model 1. The zone of interest highlighted by LIME is located inside the biopsy. However, although it focuses somehow in the carcinoma it also takes into account large portions of benign tissue and stroma. Therefore, the pathologist graded this image with a ``3'' following the evaluation rubric in Table \ref{tab:rubric}. Model 2 and 5 obtained worse results as the zone of interest highlighted by LIME is mostly located outside the biopsy. For this reason, the pathologist graded these two samples with a ``1''.

\begin{figure}[htbp!]
    \centering
    \begin{subfigure}{0.4\textwidth}
        \includegraphics[width=\textwidth]{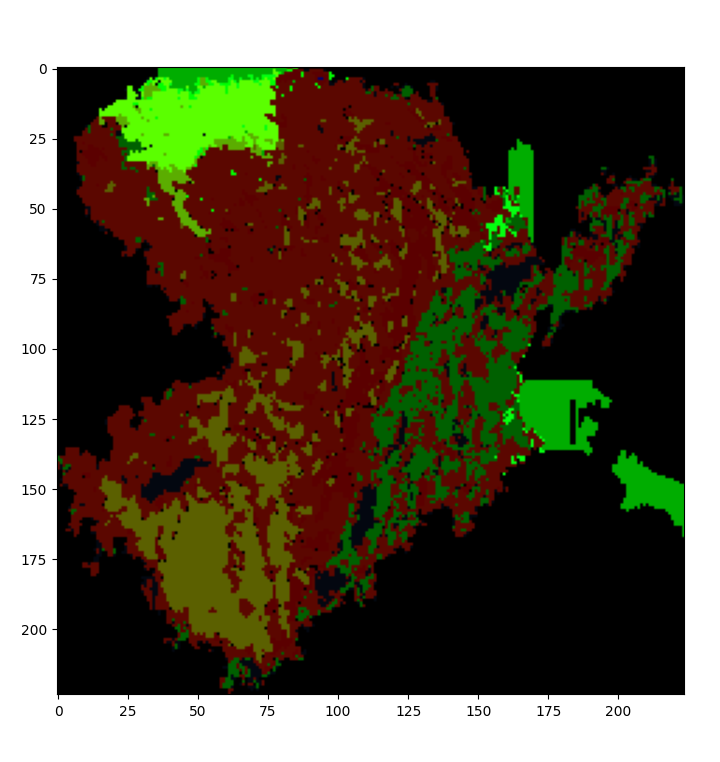}
    \end{subfigure}
    \hfill
    \begin{subfigure}{0.4\textwidth}
        \includegraphics[width=\textwidth]{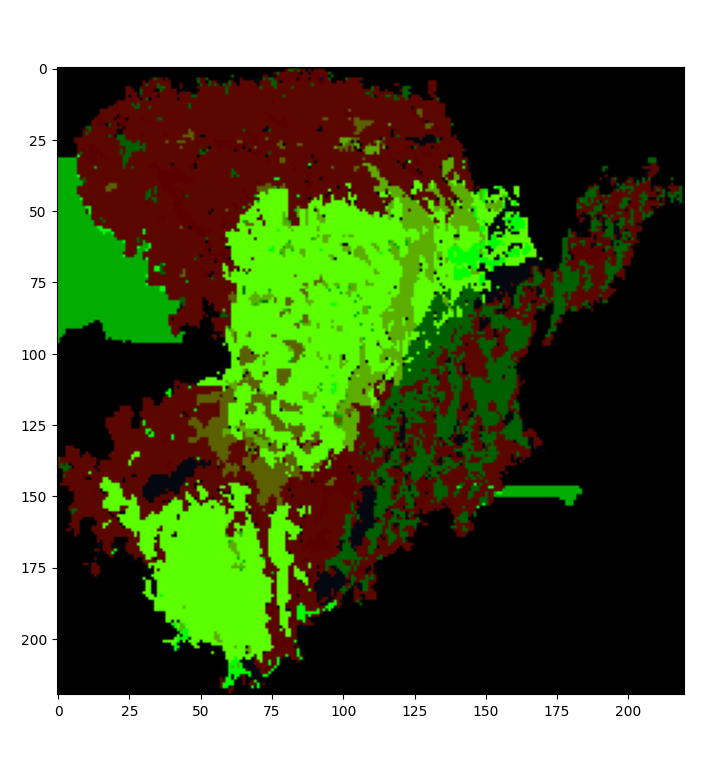}
    \end{subfigure}

    \begin{subfigure}{0.4\textwidth}
        \includegraphics[width=\textwidth]{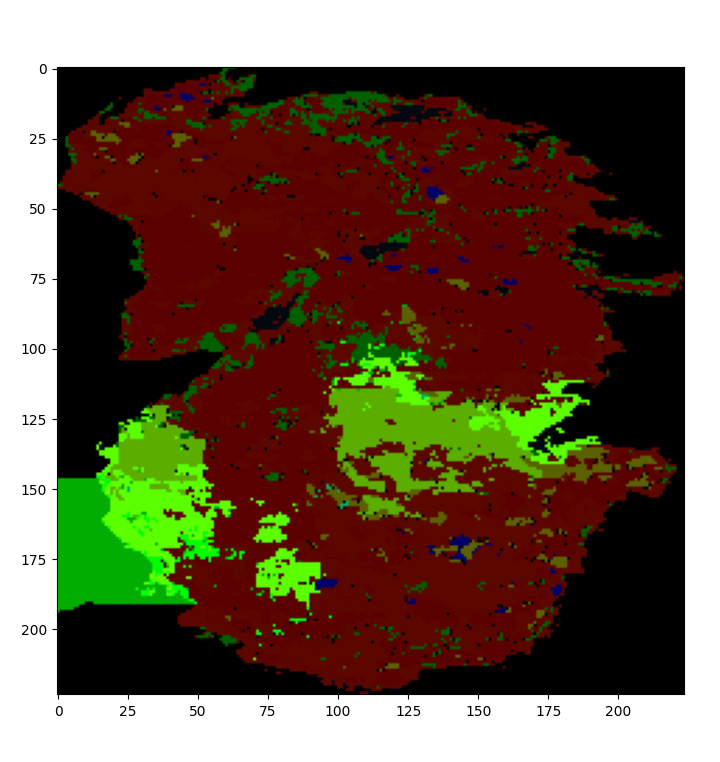}
    \end{subfigure}
    \hfill
    \begin{subfigure}{0.4\textwidth}
        \includegraphics[width=\textwidth]{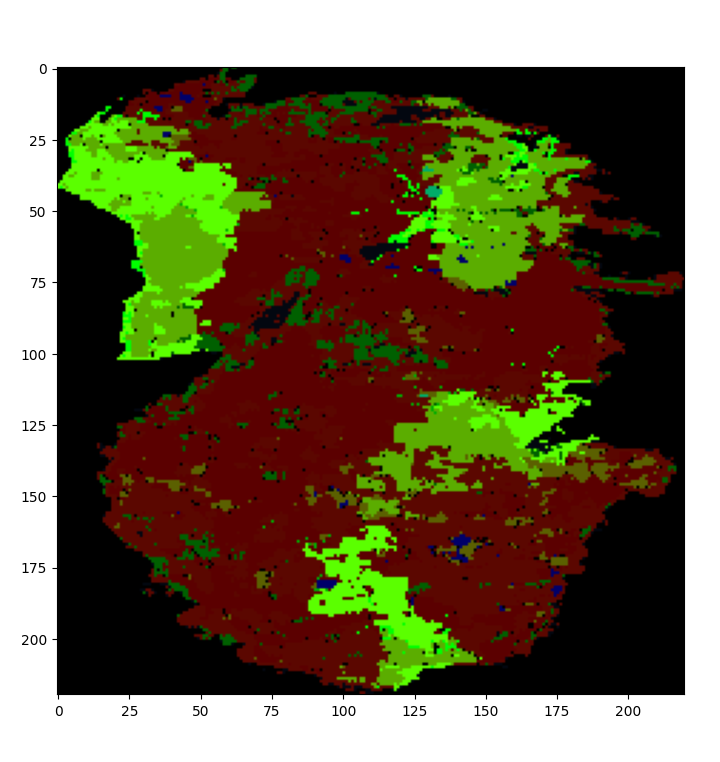}
    \end{subfigure}

    \begin{subfigure}{0.4\textwidth}
        \includegraphics[width=\textwidth]{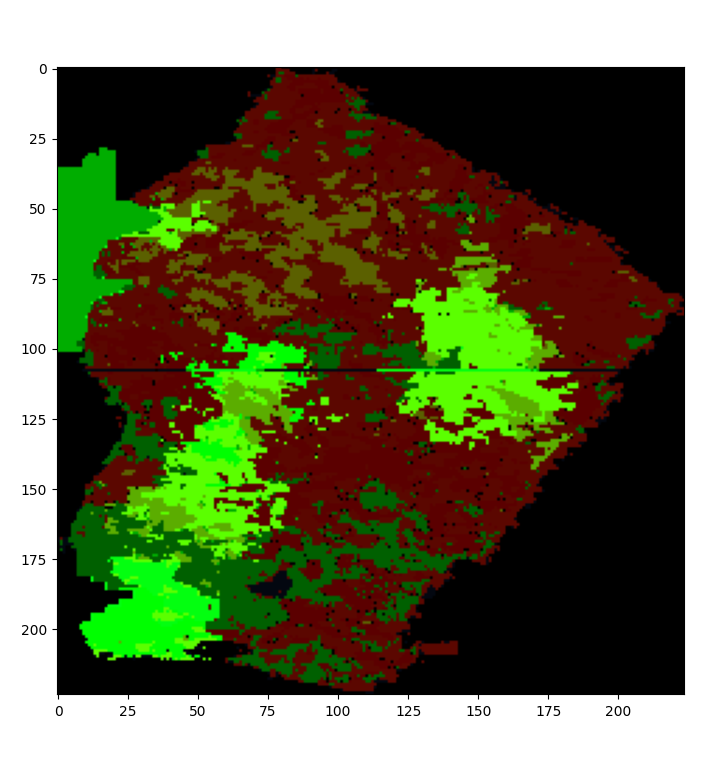}
    \end{subfigure}
    \hfill
    \begin{subfigure}{0.4\textwidth}
        \includegraphics[width=\textwidth]{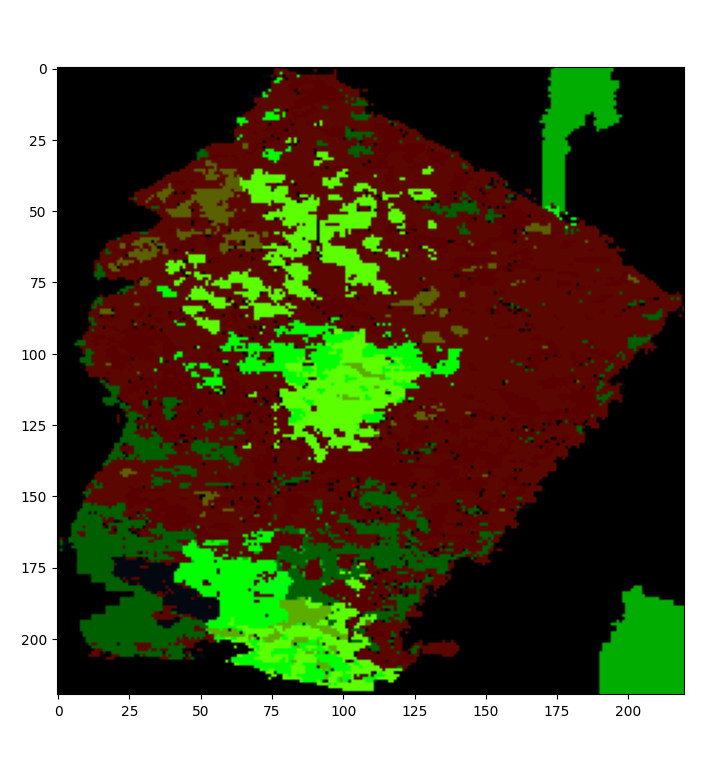}
    \end{subfigure}    
    
    \caption{Explainability results of the original model (left column) and the results of the optimized model (right column) for three diffent type of cancer: Her2 (top row), Luminal A (center row) and Luminal B (bottom row)}
    \label{fig:images_original_optimized}
\end{figure}

At the end of the optimization process we obtained a model whose interpretation capabilities were improved in comparison with the original model. In Figure \ref{fig:images_original_optimized} we can see an example of explainability results of the original model (left column) and the results of the optimized model (right column). As can be seen, the optimized model shows better results in terms of the interpretation area.

\section{Discussion and conclusions}\label{sec:discussion}

In this work we have seen several areas where Human-in-the-Loop (HITL) strategies can be useful when training machine learning models. First of all we can say that the medical domain is especially suitable for this type of techniques, since the datasets are full of uncertainty and incompleteness, and the problems to solve are hard.

Deep learning (DL) models are well suited to work with unstructured data (images, video, signals, etc.) because they are able to learn the complex patterns in the data by working directly with raw data. They do this through an automatic feature engineering process that is part of the learning process. This allow the ML engineers not to rely on  handcrafting features, but on operating on a mathematical entity so that it discovers representations from the training data autonomously. 

But this DL approach has an important drawback, it requires large amounts of data to be able to infer models. In medicine we can find big databases with clinical cases but when we focus on a particular problem (e.g. breast cancer), the data available may not be sufficient to train a deep learning model. A \textit{doctor-in-the-loop} approach can use human expertise and long-term experience to fill the gaps in large amounts of data or deal with complex data.

Regarding the segmentation process, we can see that a HITL technique such as Interactive Machine Learning (IML) is particularly suitable in the medical domain because IML seems to be especially useful in giving additional structure to something that does not have it (images). It should not be forgotten that humans are also good at giving structure to data that do not have structure. This interactive image segmentation process is an important tool in several fields such as biomedical imaging, material science, geology, manufacturing, etc.

In our case, the pathologist's involvement allowed us to pay more attention to the cancerous areas in the image and identify them with maximum accuracy by trying to group areas and simplifying the information presented to make it more useful for further diagnosis. The main problem of segmentation came from the identification of the normal adipose tissue and normal ductal/glandular structures, but these parts were not considered as important for subsequent cancer classification.

Regarding the classification task, our work tried to apply HITL deep-learning methods for integration of genomic data and WSI-based analysis. This is an open problem in this field and the results obtained by the models were not optimal. This is due to several reasons: it is a complex problem that is still under research and the data available for certain types of cancer were not enough for the models to be able to generalize patterns.

In this case the pathologist could not help us beyond trying to improve the segmentation in terms of better detection of the cancer areas, to try to make the models more focused on those parts at the time of classification, but the current medical knowledge did not allow us to go further. 

Since the models used for classification were not offering optimal results we decide to use the feedback from the pathologist to, at least, try to improve the interpretability results from those models carrying out a HITL hyperparameter optimization process.

During the interpretability analysis process we use post-hoc model-specific interpretability methods---which look inside a non-transparent model to try to draw conclusions about its performance---and model-agnostic---which treat the model as a black box and draw their conclusions by varying the inputs and seeing how these changes affect the outputs---. Although at first it might seem that model-specific methods would have more information about the model and thus provide better explainability than model-agnostic ones, we see that this is not the case and the best results are obtained with a model-agnostic method as LIME.

Also this variability obtained with the current post-hoc interpretability methods raises the question to what extent these algorithms are really reliable and really indicate what the ML model is doing. Some authors pinpoint that it is necessary to develop some way of validating the explanatory capabilities obtained by these models, especially in complex domains, such as medical environments, where it may not be easy to determine the correct answer to a given problem. A HITL approach can be a mean to perform such evaluation.

In our case the variations in interpretability may also be a reflection of the low accuracy of the models, and it is to be expected that with better agreement metrics the explainability algorithms will be more consistent in their responses.

As a final conclusion we can say that our contribution demonstrates how a human-in-the-loop process improved certain aspects of machine learning model building. We also highlight the limitations of this approach: despite involving human experts, complex domains can still pose challenges, and collaboration may not always be effective.

In any case, the expert helped with his feedback to select those models that presented better explainability, even if their accuracy was the same. This allows us to focus on those models that lead us to the right answers for the right reasons (and not taking into account spurious information or correlations).

As future work, we plan to approach the problem from the genomical point of view. Pathologists usually use 50 well-known genes to the detection of cancer, but humans have around 20000 genes that can be analyzed. Machine learning algorithms can be used to find a new subset of genes to improve the breast cancer detection. 

\backmatter

\bmhead{Acknowledgments}

This work has been supported by the State Research Agency of the Spanish Government (Grant PID2019-107194GB-I00/AEI/10.13039/501100011033) and by the Xunta de Galicia (Grant ED431C 2022/44), supported by the EU European Regional Development Fund (ERDF). We wish to acknowledge support received from the Centro de Investigación de Galicia CITIC, funded by the Xunta de Galicia and ERDF (Grant ED431G 2019/01). The results published here are in whole or part based upon data generated by the TCGA Research Network: https://www.cancer.gov/tcga





\bibliography{BRCA-HITL-bibliography}


\end{document}